\documentclass[review]{elsarticle}
\usepackage{color}
\usepackage{lineno,hyperref}
\usepackage{amsmath}
\usepackage{algorithm}
\usepackage{algorithmic}
\usepackage{multirow}
\modulolinenumbers[5]

\journal{Journal of \LaTeX\ Templates}

\begin{document}

\begin{frontmatter}

\title{Localization-aware Channel Pruning for Object Detection }
%\tnotetext[mytitlenote]{Fully documented templates are available in the elsarticle package on %\href{http://www.ctan.org/tex-archive/macros/latex/contrib/elsarticle}{CTAN}.}

\author[mymainaddress]{Zihao Xie}
\ead{zihaoxie@hust.edu.cn}
%\address{School of Artificial Intelligence and Automation, Huazhong University of Science and Technology, Wuhan 430074, China}
\author[mymainaddress]{Li Zhu\corref{mycorrespondingauthor}}
\cortext[mycorrespondingauthor]{Corresponding author}
\ead{lizhu2016@hust.edu.cn}
\author[mymainaddress]{Lin Zhao}


\author[mysecondaryaddress]{Bo Tao}

\author[mythirdaddress]{Liman Liu}

\author[mymainaddress,mysecondaddress]{Wenbing Tao}

\ead{wenbingtao@hust.edu.cn}

\address[mymainaddress]{National Key Laboratory of Science and Technology on Multi-spectral Information Processing, School of Artificial Intelligence and Automation, Huazhong University of Science and Technology, Wuhan 430074, China}
\address[mysecondaddress]{Shenzhen Huazhong University of Science and Technology Research Institute, Shenzhen, 518057, China}
\address[mysecondaryaddress]{State Key Laboratory of Digital Manufacturing Equipment and Technology, School of Mechanical Science and Engineering, Huazhong University of Science and Technology, Wuhan, Hubei 430074, PR China}
\address[mythirdaddress]{School of Biomedical Engineering, South-Central University for Nationalities, Wuhan 430074, China}
%\address[myfourthaddress]{ School of Mathematics and Statistics, Xidian University, Xi’an 710126, China }
%{jyuan,xcfeng}@xidian.edu.cn2 




\begin{abstract}
Channel pruning is one of the important methods for deep model compression. Most of existing pruning methods mainly focus on classification. Few of them conduct systematic research on object detection. However, object detection is different from classification, which requires not only semantic information but also localization information. In this paper, based on \textcolor[rgb]{0,0,1}{discrimination-aware channel pruning (DCP)} which is state-of-the-art pruning method for classification, we propose a localization-aware auxiliary network to find out the channels with key information for classification and regression so that we can conduct channel pruning directly for object detection, which saves lots of time and computing resources. In order to capture the localization information, we first design the auxiliary network with a contextual RoIAlign layer which can obtain precise localization information of the default boxes by pixel alignment and enlarges the receptive fields of the default boxes when pruning shallow layers. Then, we construct a loss function for object detection task which tends to keep the channels that contain the key information for classification and regression. Extensive experiments demonstrate the effectiveness of our method. On MS COCO, we prune 70\% parameters of the SSD based on ResNet-50 with modest accuracy drop, which outperforms the-state-of-art method. 
\end{abstract}


\end{frontmatter}

%\linenumbers

\section{Introduction}

Since AlexNet \cite{krizhevsky2012imagenet} won the ImageNet Challenge: ILSVRC 2012 \cite{russakovsky2015imagenet}, deep convolutional neural network (CNNs) have been widely applied to various computer vision tasks, from basic image classification tasks \cite{he2016deep} to some more advanced applications, e.g., object detection \cite{liu2016ssd,ren2015faster}, semantic segmentation \cite{noh2015learning}, video analysis \cite{wang2016temporal} and many others. In these fields, CNNs have achieved state-of-the-art performance compared with traditional methods based on manually designed visual features.

However, deep models often have a huge number of parameters and its size is very large, which incurs not only huge memory requirement but also unbearable computation burden. As a result, a typical deep model is hard to be deployed on resource constrained devices, e.g., mobile phones or embedded gadgets. To make CNNs available on resource-constrained devices, there are lots of studies on model compression, which aims to reduce the model redundancy without significant degeneration in performance. Channel pruning \cite{he2017channel,luo2017thinet,jiang2018efficient} is one of the important methods. Different from simply making sparse connections \cite{han2015deep,han2015learning}, channel pruning reduces the model size by directly removing redundant channels and can achieve fast inference without special software or hardware implementation.

In order to determine which channels to reserve, existing reconstruction-based methods \cite{he2017channel,luo2017thinet,jiang2018efficient} usually minimize the reconstruction error of feature maps between the original model and the pruned one. However, a well-reconstructed feature map may not be optimal for  there is a gap between intermediate feature map and the performance of final output. Information redundancy channels could be mistakenly kept to minimize the reconstruction error of feature maps. To find the channels with true discriminative power for the network, DCP \cite{zhuang2018discrimination} attend to conduct channel selection by introducing additional discrimination-aware losses that are actually correlated with the final performance. It constructs the discrimination-aware losses by a fully connected layer which works on the entire feature map. However, the discrimination-aware loss of DCP is designed for classification task. Since object detection network uses the classification network as backbone, a simple method to conduct DCP for object detection is to fine-tune the pruned model, which was trained on classification dataset, for the object detection task. But the information that the two tasks need is not exactly the same. The classification task needs strong semantic information while what the object detection task needs is not only semantic information but also localization information. Hence, the existing training scheme may not be optimal due to the mismatched goals of feature learning for classification and object detection task.

In this paper, we propose a method called localization-aware channel pruning (LCP), which conducts channel pruning directly for object detection. We propose a localization-aware auxiliary network for object detection task. First, we design the auxiliary network with a contextual \textcolor[rgb]{0,0,1}{RoIAlign} layer which can obtain precise localization information of the default boxes by pixel alignment and enlarges the receptive fields of the default boxes when pruning shallow layers. Then, we construct a loss function for object detection task which tends to keep the channels that contain the key information for classification and regression. Our main contributions are summarized as follows. (1) We propose a localization-aware auxiliary network which can find out the channels with key information so that we can conduct channel pruning directly on object detecion dataset, which saves lots of time and computing resources. (2) We propose a contextual RoIAlign layer which enlarges the receptive fields of the default boxes in shallow layers. (3) Extensive experiments on benchmark datasets show that the proposed method is theoretically reasonable and practically effective. For example, our method can prune 70\% parameters of SSD \cite{liu2016ssd}  based on ResNet-50 \cite{he2016deep} with modest accuracy drop on VOC2007, which outperforms the-state-of-art method.
\section{Related Works}

\subsection{Network Quantization}
Network quantization compresses the original network by reducing the number of bits required to represent each weight. Han et al. \cite{han2015deep} propose a complete deep network compression pipeline: First trim the unimportant connections and retrain the sparsely connected network. Weight sharing is then used to quantize the weight of the connection, and then the quantized weight and codebook are Huffman encoded to further reduce the compression ratio. Courbariaux et al. \cite{courbariaux2016binarized} propose to accelerate the model by reducing the weight and accuracy of the output, because this will greatly reduce the memory size and access times of the network, and replace the arithmetic operator with a bit-wise operator. Li et al. \cite{li2016ternary}  consider that multi-weights have better generalization capabilities than binarization and the distribution of weights is close to a combination of a normal distribution and a uniform distribution.  Zhou et al. \cite{zhou2017incremental} propose a method which can convert the full-precision CNN into a low-precision network, making the weights 0 or 2 without loss or even higher precision (shifting can be performed on embedded devices such as FPGAs). \textcolor[rgb]{0,0,1}{For more recent works, Yu et al. \cite{yu2019double}, to reduce the communication complexity, propose a general scheme for quantizing both model parameters and gradients.  Zhao et al. \cite{zhao2019focused} attend to the statistical properties of sparse CNNs and present focused quantization, a novel quantization strategy based on power-of-two values, which exploits the weight distributions after fine-grained pruning.}

\subsection{Sparse or Low-rank Connections}
Wen et al. \cite{wen2016learning} propose a learning method called Structured Sparsity Learning, which can learn a sparse structure to reduce computational cost, and the learned structural sparseness can be effectively accelerate for hardware. Guo et al. \cite{guo2016dynamic} propose a new network compression method, called dynamic network surgery, is  to reduce network complexity through dynamic connection pruning. Unlike previous methods of greedy pruning, this approach integrates join stitching throughout the process to avoid incorrect trimming and maintenance of the network. Jin et al. \cite{jin2016training} proposes to reduce the computational complexity of the model by training a sparsely high network. By adding a $l_{0}$ paradigm about weights to the loss function of the network, the sparsity of weights can be reduced. \textcolor[rgb]{0,0,1}{For more recent works, Kim \cite{Kim_2019_CVPR} propose novel accuracy metrics to represent the accuracy and complexity relationship for a given neural network and  use these metrics in a non-iterative fashion to obtain the right rank configuration which satisfies the constraints on FLOPs and memory while maintaining sufficient accuracy. Liu et al. \cite{liulayerwise} propose a layerwise sparse coding (LSC) method to maximize the compression ratio by extremely reducing the amount of meta-data.}

\subsection{Channel Pruning}
Finding unimportant weights in the network has a long history. LeCun \cite{lecun1990optimal} and Hassibi \cite{hassibi1993second} consider using the Hessian, which contains second order derivative, performs better than using the magnitude of the weights. Computing the Hessian is expensive and thus is not widely used. Han \cite{han2015deep} et al. proposed an iterative pruning method to remove the redundancy in deep models. Their main insight is that small-weight connectivity below a threshold should be discarded. In practice, this can be aided by applying $l_{1}$ or $l_{2}$ regularization to push connectivity values to become smaller. The major weakness of this strategy is the loss of universality and flexibility, thus seems to be less practical in real applications. Li et al. \cite{li2016pruning} measure the importance of channels by calculating the sum of absolute values of weights. Hu et al. \cite{hu2016network} define average percentage of zeros (APoZ) to measure the activation of neurons. Neurons with higher values of APoZ are considered more redundant in the network. With a sparsity regularizer in the objective function \cite{alvarez2016learning,liu2017learning}, training based methods are proposed to learn the compact models in the training phase. With the consideration of efficiency, reconstruction-methods \cite{he2017channel,luo2017thinet} transform the channel selection problem into the optimization of reconstruction error and solve it by a greedy algorithm or LASSO regression. DCP \cite{zhuang2018discrimination} aimed at selecting the most discriminative channels for each layer by considering both the reconstruction error and the discrimination-aware loss. {PP \cite{singh2019play} jointly prunes and fine-tunes CNN model parameters, with an adaptive pruning rate. Liu et al. \cite{liu2019metapruning} propose a novel meta learning approach for automatic channel pruning of very deep neural networks. AutoPrune \cite{xiao2019autoprune} prunes the network through optimizing a set of trainable auxiliary parameters instead of original weights.}

\subsection{Object Detection}
\textcolor[rgb]{0,0,1}{Current state-of-the-art object detectors with deep learning can be mainly divided into two major categories: two-stage detectors and one-stage detectors. Two-stage detectors first generate region proposals which may potentially
be objects and then make predictions for these proposals. Faster R-CNN \cite{ren2015faster} is a representative two-stage detector, which was able to make predictions at 5FPS on GPU and achieved state-of-the-art results on many public benchmark datasets, such as Pascal VOC 2007, 2012 and MSCOCO. Currently, there are huge number of detector variants based on Faster R-CNN for different usage \cite{lin2017feature, cai2018cascade}. Mask R-CNN \cite{he2017mask} extends Faster R-CNN to the field of instance segmentation. Based on Mask R-CNN, Huang et al. \cite{huang2019mask} proposed a mask-quality aware framework, named Mask Scoring R-CNN, which learned the quality of the predicted masks and calibrated the misalignment between mask quality and mask confidence score. }

\textcolor[rgb]{0,0,1}{Different from two-stage detectors, one-stage detectors do not generate proposals and directly make predictions on the whole feature map. SSD \cite{liu2016ssd} is a representative one-stage detector. However, the class imbalance between foreground and background is a severe problem in one-stage detector. RetinaNet \cite{lin2017focal} matigates the class imbalance problem by introducing focal loss which reduces loss of easy samples. The previous approaches required designing anchor boxes manually to train a detector. Currently, a series of anchor-free object detectors were developed, where the goal was to predict keypoints of the bounding box, instead of trying to fit an object to an anchor. Law and Deng proposed a novel anchor-free framework CornerNet \cite{law2018cornernet} which detected objects as a pair of corners.  Later there were several other variants of anchor-free detectors \cite{zhou2019objects, duan2019centernet}}

\section{Proposed Method}
\begin{figure*}[t]
	\begin{center}
		\includegraphics[width=1\linewidth]{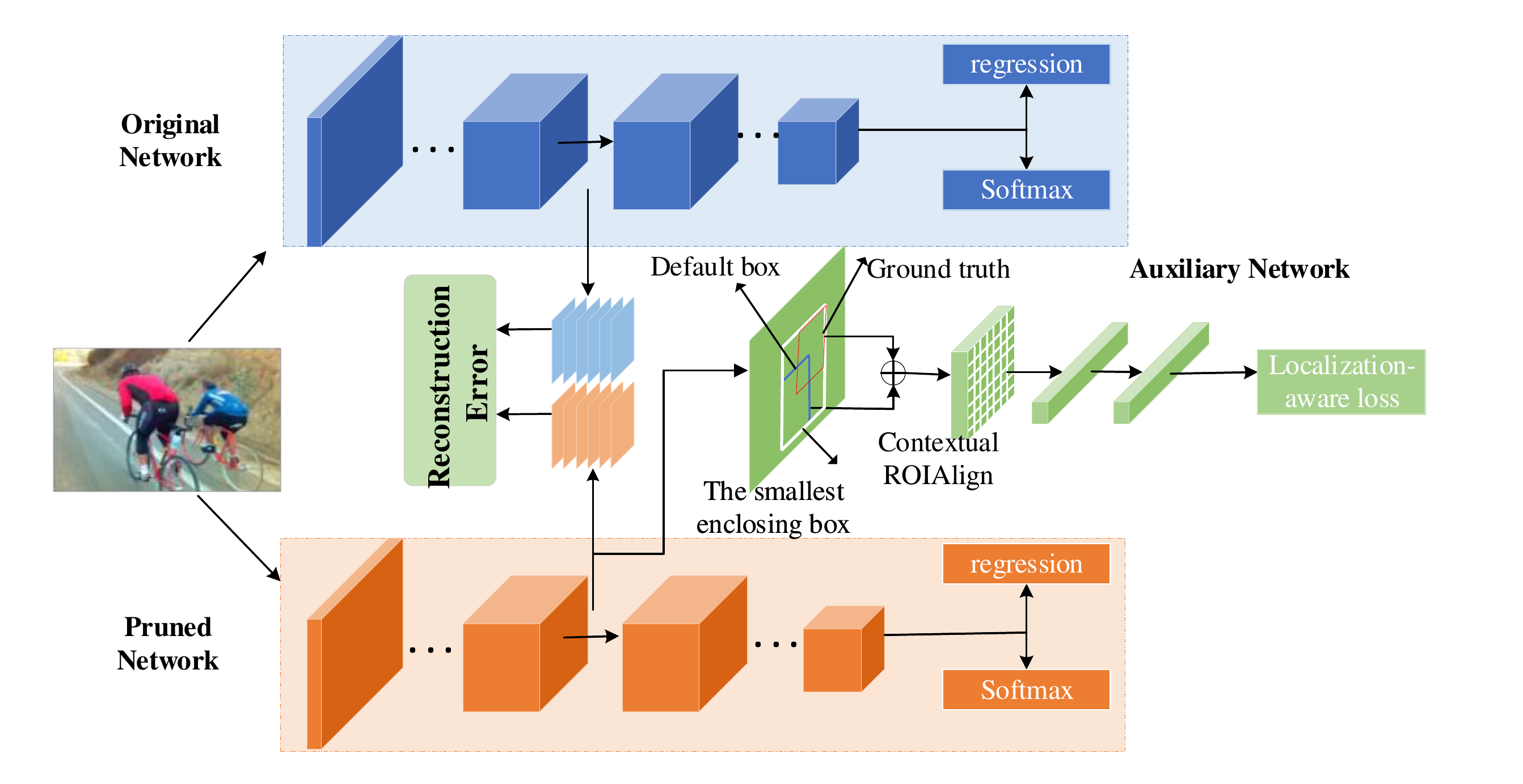}
	\end{center}
	\caption{Illustration of localization-aware channel pruning. The auxiliary network is used to supervise layer-wise channel selection. The blue network the original model, the orange network is the model to be pruned, the green network is the auxiliary network. The reconstruction error and the auxiliary network are used to construct localization-aware loss.}
	\label{LCP}
\end{figure*}

\textcolor[rgb]{0,0,1}{Fig. \ref{LCP} is the overall frame diagram. The blue network in Figure 1 is the original model which could be any detectors, such as SSD \cite{liu2016ssd}, Faster R-CNN \cite{ren2015faster} and so on. The orange network in Figure 1 is the model to be pruned which is initialized exactly the same as the original model. The middle is the auxiliary network. We use the auxiliary network to prune the model by constructing the localization-aware loss. The localization-aware loss consists of two parts, one part is the reconstruction error and the other is the loss of auxiliary network. Their details will be discussed later. After the loss is constructed, we could fine-tune the network and use the gradient of the localization-aware loss to decide which channel to preserve. After repeating this operation layer by layer, the pruning is finished.}

The auxiliary network we propose mainly consists of two parts. First, a contextual RoIAlign layer is designed to extract the features of the boxes. Then, a loss is designed for object detection task which can reserve the important channels. The details of the proposed approach are elaborated below.
\subsection{Contextual RoIAlign Layer}
\begin{figure}[t]
	\begin{center} 
		\includegraphics[width=1.0\linewidth]{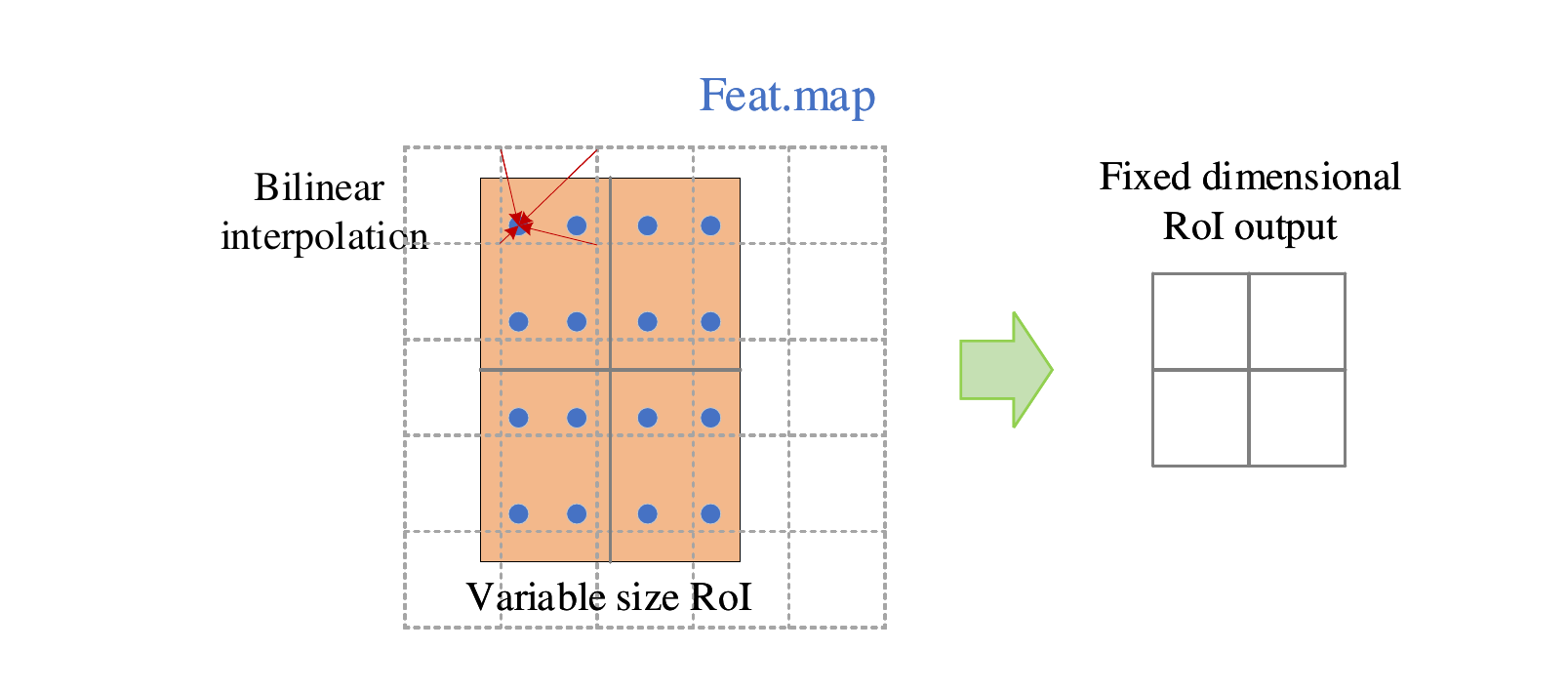}
	\end{center}
	\caption{RoIAlign: The dashed grid represents a feature map, the solid lines an RoI (with 2$\times$ bins in this example), and the dots the 4 sampling points in each bin. RoIAlign computes the value of each sampling point by bilinear interpolation from the nearby grid points on the feature map.}
	\label{roialign}
\end{figure}

\begin{figure}[t]
	\begin{center} 
		\includegraphics[width=1.0\linewidth]{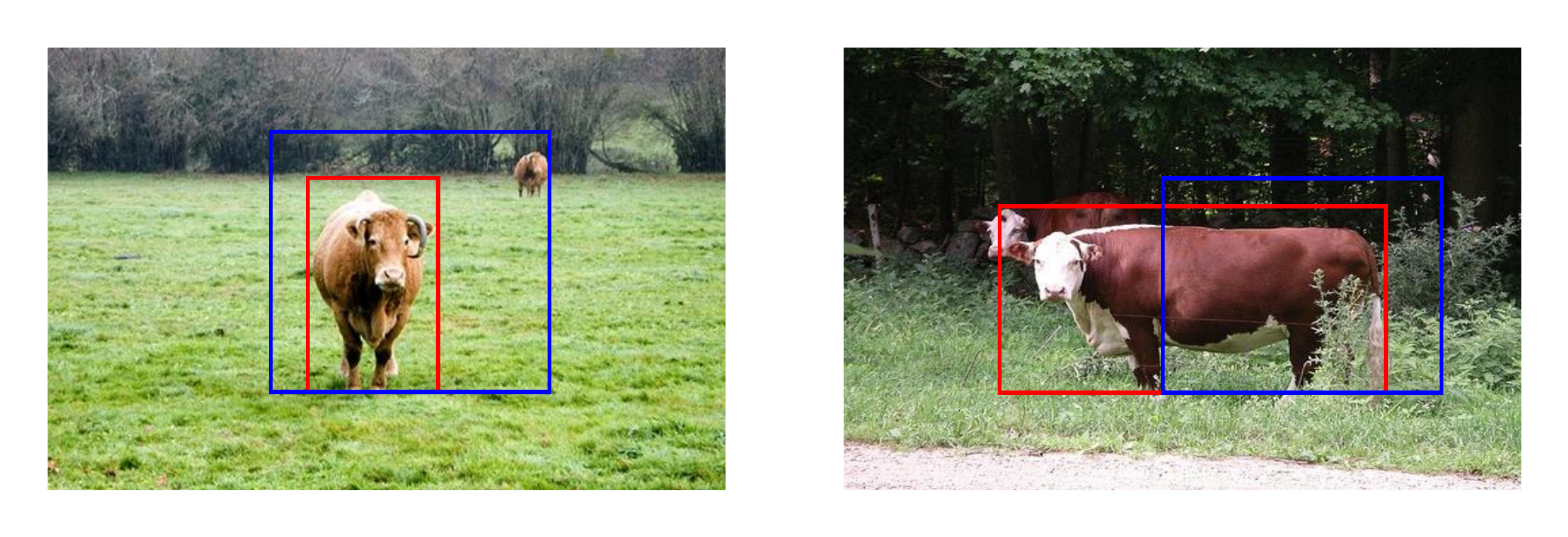}
	\end{center}
	\caption{The features of default boxes do not always contain enough context information, especially when we prune shallow layers. The blue is default box, the red is ground truth.}
	\label{proposal}
\end{figure}

For object detection task, if we predict the bounding boxes directly on the entire feature maps, there will be a huge amount of parameters and unnecessary noises. So, it is important to extract the feature of region of interest (RoI) , which can be better used for classification and regression. To obtain precise localization information and find out the channels which are important for classification and regression, RoIAlign layer is a good choice which properly align the extracted features with the input. RoIAlign use bilinear interpolation to compute the exact values of the input features at four regularly sampled locations in each RoI bin, and aggregate the result (using max or average), see Fig. \ref{roialign} for details. However, the default boxes generated by the detector do not always completely cover the object area. From Fig. \ref{proposal}, we can see that, the defalut box is sometimes bigger than the ground truth and sometimes smaller than it. So, the receptive fields may be insufficient if we only extract the features of the default box especially when we prune the shallow layers. To solve this problem, we propose a contextual RoIAlign layer, which introduces larger context information. \textcolor[rgb]{0,0,1}{The orange part in Figure 4 is feature extraction network. The network first extract the feature map of the whole image, then obtain the features of the boxes by RoIAlign operation. To introduce larger context information, we further gather the information of the default box and its context by adding the feature of the default box and its enclosing convex object.}

\begin{figure}[t]
	\begin{center} 
		\includegraphics[width=1.0\linewidth]{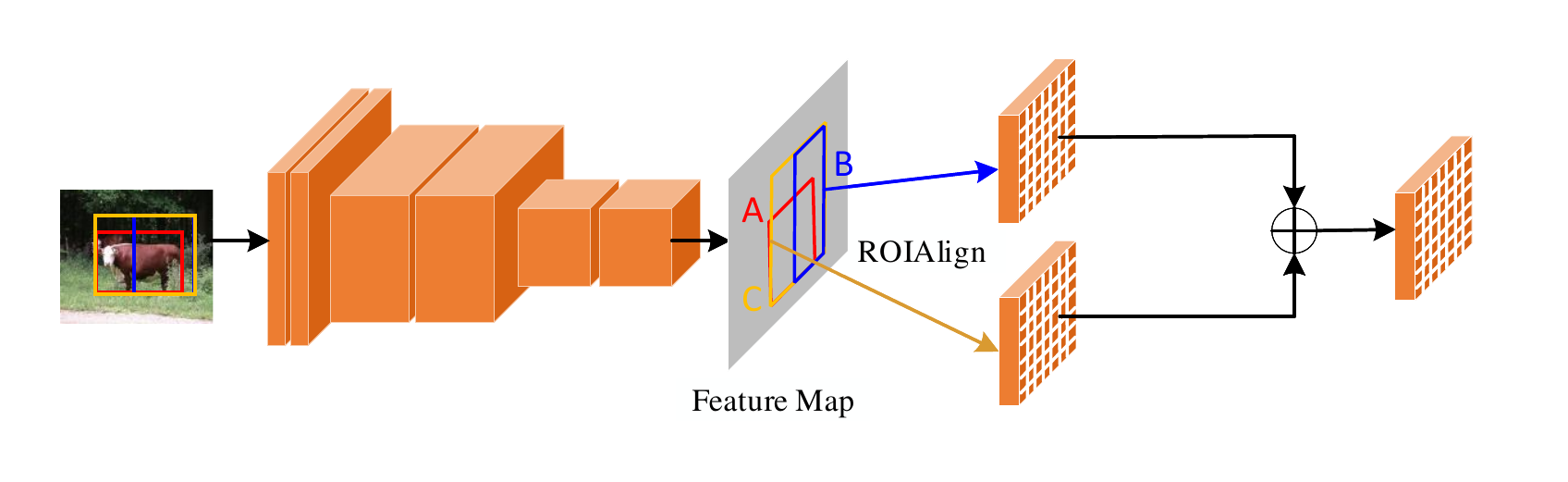}
	\end{center}
	\caption{Contextual RoIAlign: The red, blue and orange boxes represent the ground truth, default box and its smallest enclosing box, respectively. The network first extract the feature map of the whole image, then obtain the features of the boxes by RoIAlign operation. To introduce larger context information, we further gather the information of the default box and its context by adding the feature of the default box and its enclosing convex object.
	}
	\label{cr}
\end{figure}

For better description of the algorithm, some notations are given first. For a training sample, $( x_{a1}, y_{a1}, x_{a2}, y_{a2})$ represents the coordinates of ground truth box $A$, $( x_{b1}, y_{b1}, x_{b2}, y_{b2})$ denotes the coordinates of the matched default box $B$. We further use $\mathcal{F}$ to denote the feature map and $\mathcal{F_{S}}$ represents the features of area $S$, $RoIAlign$ represents the RoIAlign operation. First, we calculate the $IoU$ of box A and B:
\begin{align}\label{eq1}
IoU_{AB}=\frac{A \cap B}{A \cup B}
\end{align}
B is a positive sample only if $IoU_{AB}$ is larger than a preset threshold. We do not conduct contextual RoIAlign for B when B is negative sample. If B is a positive sample, then we calculate the smallest enclosing convex object C for A and B:
\begin{align}\label{eq2}
x_{c1}= min(x_{a1},x_{b1})
\end{align}
\begin{align}\label{eq3}
y_{c1}= min(y_{a1},y_{b1})
\end{align}
\begin{align}\label{eq4}
x_{c2}= max(x_{a2},x_{b2})
\end{align}
\begin{align}\label{eq5}
y_{c2}= max(y_{a2},y_{b2})
\end{align}
where $( x_{c1},y_{c1},x_{c2},y_{c2})$ are the coordinates of C. Finally, the output of contextual RoIAlign layer is defined as:
\begin{align}\label{eq6}
\mathcal{F_{O}} = RoIAlign(\mathcal{F_{B}})+RoIAlign(\mathcal{F_{C}})
\end{align}
Now we can get the precise features of default box B, the process can refer to Fig \ref{cr}.
\subsection{Construction of the Loss of Auxiliary Network}
After we construct the contextual RoIAlign layer, we need to consturct a loss for object detection task so that we can use the gradient of the auxiliary network to conduct model pruning. The details are discussed below.

\textcolor[rgb]{0,0,1}{Considering that the object detection task needs both the classification and localization information. The loss of auxiliary consists of two parts. One is the loss for classification and the other is the loss for regression. For the classification part, We still use discrimination-aware loss which has been proved to be useful on classification task. For the regression part, we choose GIoU \cite{rezatofighi2019generalized} loss function.} It is reasonable to use GIoU as loss function for boxes regression. It considers not only overlapping areas but also non-overlapping areas, which better reflects the overlap of the boxes. The GIoU of two arbitrary shapes $A$ and $B$ is defined as:
\begin{figure}[t]
	\begin{center}
		\includegraphics[width=0.5\linewidth]{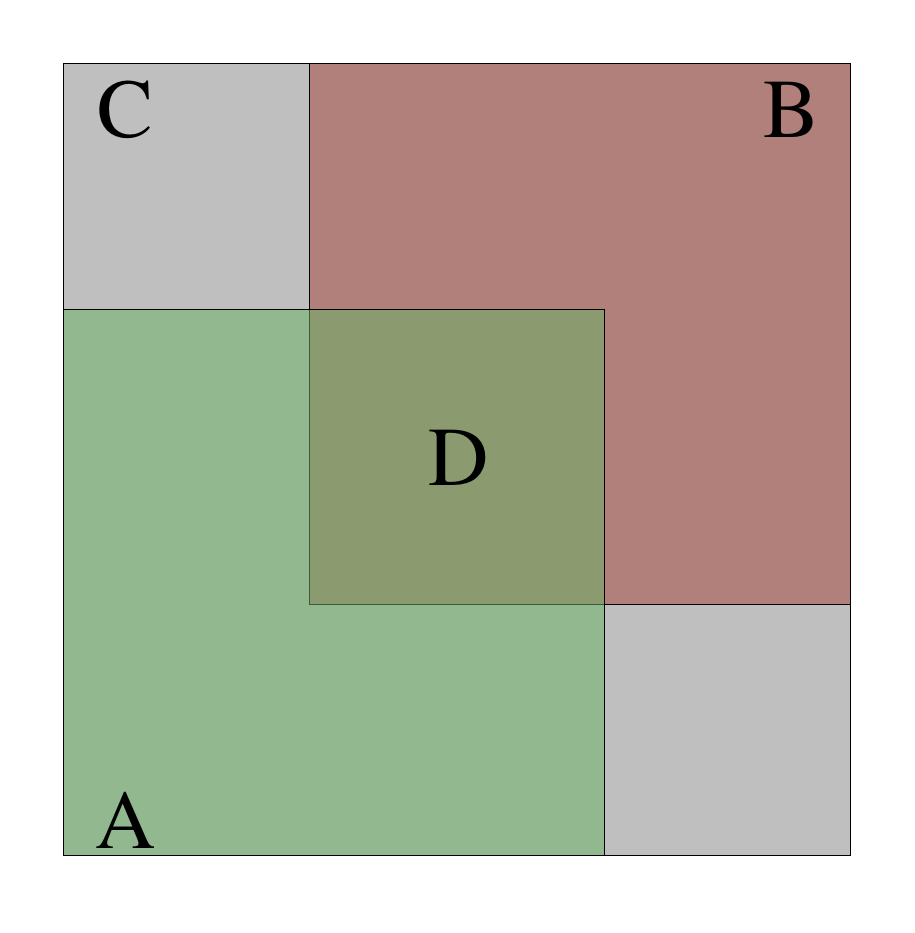}
	\end{center}
	\caption{A, B are two arbitrary shapes, C is the smallest enclosing convex of A and B, D is the $[IoU]$ of A and B. }
	\label{GIOU}
\end{figure}
\begin{align}\label{eq9}
GIoU_{AB}=IoU_{AB}-\frac{C-U}{C}
\end{align}
where $U=A+B-IoU_{AB}$, $IoU_{AB}$ and $C$ are calculated by Eq. \ref{eq1} - Eq. \ref{eq5}. Fig. \ref{GIOU} is a schematic diagram of GIoU. Then, we use $G_{i}$ to denote the GIoU of the $i$-th predicted box and the ground truth, $E_{i}$ to represent the cross entropy of the $i$-th predicted box. Then, in the pruning stage, $\mathcal{L}_{ac}$ represents the classification loss, $\mathcal{L}_{ar}$ represents the regression loss, $\mathcal{L}_{a}$ represents the localization-aware loss of the auxiliary network. Finally, the loss of positive samples in pruning stage is defined as:
\begin{align}\label{eq10}
\mathcal{L}_{ac}=\sum_{i}E_{i}
\end{align}
\begin{align}\label{eq11}
\mathcal{L}_{ar}=\sum_{i}m (1-G_{i})
\end{align}
\begin{align}\label{eq12}
\mathcal{L}_{a}=\mathcal{L}_{ac}+\mathcal{L}_{ar}
\end{align}
where $m$ is a constant coefficient.
\subsection{Localization-aware Channel Pruning}
After we construct the auxiliary network and the localization-aware loss, we can conduct channel pruning with them layer by layer. The pruning process of the whole model is described in Algorithm \ref{pruning}. For better description of the channel selection algorithm, some notations are given first. Considering a $L$ layers of the CNN model and we are pruning the $l$-th layer, $X$ represents the output feature map of the $lth$ layer, $W$ denotes the convolution filter of the $(l+1)$-th layer of the pruned model and $*$ represents the convolution operation. We further use  $F \in$ $R^{N \times HY}$ to denote output feature maps of the $(l+1)$-th layer of the original model. Here, $N$, $H$, $Y$ represents the number of output channels, the height and the width of the feature maps respectively. Finally we use $\mathcal{L}_{c}$ and $\mathcal{L}_{r}$ to denote classification loss and regression loss of the pruned network. 

To find out the channels which really contribute to the network, we should fine-tune the auxiliary network and pruned network first and the fine-tune loss is defined as the sum of the losses of them:
\begin{align}\label{eq14}
\mathcal{L}_{f}=\mathcal{L}_{a}  + \mathcal{L}_{c} + \mathcal{L}_{r}
\end{align}
In order to minimizing the reconstruction error of a layer, we introduce a reconstruction loss as DCP does which can be defined as the Euclidean distance of feature maps between the original model and the pruned one:
\begin{align}\label{eq15}
\mathcal{L}_{re}=\frac{1}{2Q}\Vert F - X * W_{\mathcal{C}}\Vert^{2}_{2}
\end{align}
where $Q=M \times H \times Y$, $\mathcal{C}$ represents the selected channels, $W_{\mathcal{C}}$ represents the submatrix indexed by $\mathcal{C}$.
\begin{algorithm} [h]
	\caption{The proposed method} 
	\label{pruning} 
	{\bf Input:} number of layers $L$, weights of original model $\left\{ W^{l}:0<l<L \right\}$, the training set $\left\{x_{i},y_{i}\right\}$, the pruning rate $\eta$. \\
	{\bf Output:} $\left\{ W^{l}_{\mathcal{C}}:0<l<L \right\}$: weights of the pruned model. 
	\begin{algorithmic}[1] 
		\STATE Initialize $W^{l}_{\mathcal{C}}$ with $W^{l}$ for $\forall 1 \le l \le L$
		\FOR{$l = 1,2,\cdots,L$}
		\STATE Construct the fine-tune loss $\mathcal{L}_{f}$ shown as in Eq. \ref{eq14}
		\STATE Fine-tune the auxiliary network and the pruned model by  $\mathcal{L}_{f}$
		\STATE Construct the joint loss $\mathcal{L}$ shown as in Eq. \ref{eq16}
		\STATE Conduct channel selection for layer $l$ by Eq. \ref{eq17}
		\STATE Update $W^{l}_{\mathcal{C}}$ w.r.t. the selected channels by Eq. \ref{eq18}
		\ENDFOR
		\STATE return the pruned model
	\end{algorithmic} 
\end{algorithm}

Taking into account the reconstruction error, the localization-aware loss of the auxiliary network, the problem of channel pruning can be formulated to minimize the following joint loss function:
\begin{align}\label{eq16}
\nonumber \mathop{min}\limits_{W_{\mathcal{C}}} \quad   \mathcal{L}(W_{\mathcal{C}})=&\mathcal{L}_{re}(W_{\mathcal{C}})+\alpha \mathcal{L}_{a}(W_{\mathcal{C}}) \\ & s.t. \quad  \Vert \mathcal{C} \Vert_{0} \leq \mathcal{K}
\end{align}
where $\alpha$ is a constant, $\mathcal{K}$ is the number of channels to be selected. Directly optimizing Eq. \ref{eq16} is NP-hard.  Following general greedy methods in DCP, we conduct channel pruning by considering the gradient of Eq. \ref{eq16}. Specifically, the importance of the $k$-th channel is defined as:
\begin{align}\label{eq17}
\mathcal{S}_{k}=\sum_{i=1}^{H} \sum_{j=1}^{W} \Vert \frac{\partial \mathcal{L}}{\partial W_{k,i,j}} \Vert_{2}^{2}
\end{align}
\begin{figure*}[t]
	\begin{center}
		\includegraphics[width=1\linewidth]{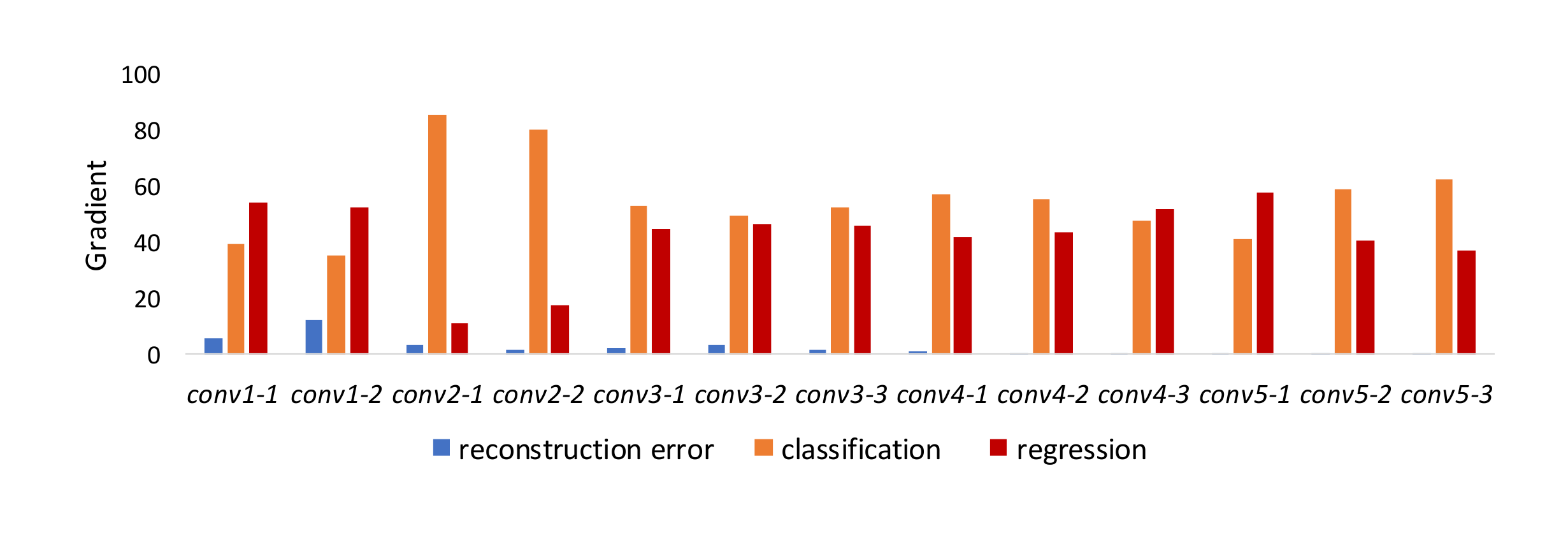}
	\end{center}
	\caption{The percentage of the gradients generated by the three loss functions.}
	\label{gradient}
\end{figure*}
$\mathcal{S}_{k}$ is the square sum of gradient of the $k$-th channel. Then we reserve the channels with the $i$ largest importance and remove others. After this, the selected channels is further optimized by stochastic gradient (SGD). $W_{\mathcal{C}}$ is updated by:
\begin{align}\label{eq18}
W_{\mathcal{C}} = W_{\mathcal{C}} - \gamma \frac{\partial \mathcal{L}}{\partial W_{\mathcal{C}}}
\end{align}
where $\gamma$ represents the learning rate. After updating $W_{\mathcal{C}}$, the channel pruning of a single layer is finished.

\section{Experiments}
We evaluate LCP on the popular 2D object detector SSD \cite{liu2016ssd}. Several state-of-the-art methods are adopted as the baselines, including \textcolor[rgb]{0,0,1}{ThiNet \cite{luo2017thinet} and DCP \cite{zhuang2018discrimination}.} In order to verify the effectiveness of our method, we use VGG and ResNet to extract feature respectively.
\subsection{Dataset and Evaluation}
The results of all baselines are reported on standard object detection benchmarks, i.e. the PASCAL VOC  \cite{everingham2010pascal} .
\textbf{PASCAL VOC2007 and 2012:} The Pascal Visual Object Classes (VOC) benchmark is one of the most widely used datasets for classification, object detection and semantic segmentation. We use the union of VOC2007 and VOC2012 trainval as training set, which contains 16551 images and objects from 20 pre-defined categories annotated with bounding boxes. And we use the VOC2007 test as test set which contains 4592 images. In order to verify the effectiveness of our method, on PASCAL VOC, we first compare our method only with ThiNet based on VGG-16 because the authors of DCP do not release the VGG model. To this end, we compare our method with DCP and ThiNet based on ResNet-50. Then we conduct the ablation experiment of our method on PASCAL VOC. In order to more fully verify the effectiveness of our method, we also perform experiments on the MS COCO2017 dataset. 

In this paper, we use $07metric$ for all experiments on PASCAL VOC. For experiments on MS COCO, the main performance measure used in this benchmark is shown by AP, which is averaging mAP across different value of IoU thresholds, i.e. $IoU=\left\{ .5,.55,\cdots,.95 \right\}$. 
\subsection{Implementation details}
Our experiments are based on SSD and the input size of the SSD is $300\times 300$. We use VGGNet and ResNet as the feature extraction network for experiments. For ThiNet, we implement it for object detection. And the three methods prune the same number of channels for each layer. Other common parameters are described in detail below. 

For VGGNet \cite{simonyan2014very}, we use VGG-16 without Batch Normalization layer and prune the SSD from conv1-1 to conv5-3. The network is fine-tuned for 10 epochs every time a layer is pruned and the learning rate is started at 0.001 and divided by 10 at epoch 5. After the model is pruned, we fine-tune it for 60k iterations and the learning rate is started at 0.0005 and divided by 10 at iteration 30k and 45k, respectively.  

For ResNet \cite{he2016deep}, we use the layers of ResNet-50 from conv1-x to conv4-x for feature extracting. The network is fine-tuned for 15 epochs every time a layer is pruned and the learning rate is started at 0.001 and divided by 10 at epoch 5 and 10, respectively. After the model is pruned, we fine-tune it for 120k iterations and the learning rate is started at 0.001 and divided by 10 at iteration 80k and 100k, respectively. 

For the loss of auxiliary network, we set $m$ to 50.

\subsection{Experiments on PASCAL VOC}
On PASCAL VOC, we prune the VGG-16 from conv1-1 to conv5-3 with compression ratio 0.75, which is 4x faster. We report the results in Tab. \ref{voc vgg}. From the results, we can see that our method achieves the best performance under the same acceleration rate. The accuracy of reconstruction based method like ThiNet drops a lot. But for our LCP, there is not much degradation in the performance of object detection. It is proved that our method retain the channels which really contribute to the final performance.  Then we conduct the experiment based ResNet-50. We report the results in Tab. \ref{voc resnet}. From the results, LCP achieves the best performance regardless of pruning by 75\% or pruning by 50\%, which proves that our method can reserve the channels which contain key information for classification and regression. In addition, the ThiNet outperforms the DCP when pruning ratio is 0.7, which indicates that pruning the model on classification dataset for object detection is not optimal.
\begin{table}[t]
	\centering
	\caption{The pruning results on PASCAL VOC2007. We conduct channel pruning from conv1-1 to conv5-3.} \label{voc vgg}
	\setlength{\tabcolsep}{1.5mm}
	\small
	\begin{tabular}{c|c|c|c|c|c}
		\hline
		\multirow{1}{*}{Method}  & \multirow{1}{*}{backbone} & \multirow{1}{*}{$\eta$}& \multirow{1}{*}{flops$\downarrow$}& \multirow{1}{*}{params$\downarrow$}& \multirow{1}{*}{mAP}\\ \hline
		\multirow{1}{*}{Original}  & \multirow{1}{*}{VGG-16} & \multirow{1}{*}{0}&\multirow{1}{*}{0}&\multirow{1}{*}{0}&\multirow{1}{*}{77.4}\\ \hline
		\multirow{1}{*}{ThiNet}  &\multirow{1}{*}{VGG-16} &\multirow{1}{*}{0.5}& \multirow{1}{*}{50\%}&\multirow{1}{*}{50\%}&\multirow{1}{*}{74.6}\\ \hline
		\multirow{1}{*}{LCP(our)} & \multirow{1}{*}{VGG-16} &\multirow{1}{*}{0.5}&\multirow{1}{*}{50\%}&\multirow{1}{*}{50\%}&\multirow{1}{*}{\textbf{77.2}}\\ \hline
		\multirow{1}{*}{ThiNet}  &\multirow{1}{*}{VGG-16} &\multirow{1}{*}{0.75}& \multirow{1}{*}{75\%}&\multirow{1}{*}{75\%}&\multirow{1}{*}{72.7}\\ \hline
		\multirow{1}{*}{LCP(our)} & \multirow{1}{*}{VGG-16} &\multirow{1}{*}{0.75}&\multirow{1}{*}{75\%}&\multirow{1}{*}{75\%}&\multirow{1}{*}{\textbf{75.2}}\\ \hline
	\end{tabular}
\end{table}
\begin{table}[t]
	\centering
	\caption{The pruning results on PASCAL VOC2007. We conduct channel pruning from conv2-x to conv4-x.} \label{voc resnet}
	\setlength{\tabcolsep}{1.5mm}
	\small
	\begin{tabular}{c|c|c|c|c|c}
		\hline
		\multirow{1}{*}{Method}  & \multirow{1}{*}{backbone} & \multirow{1}{*}{$\eta$}& \multirow{1}{*}{flops$\downarrow$}& \multirow{1}{*}{params$\downarrow$}& \multirow{1}{*}{mAP}\\ \hline
		\multirow{1}{*}{Original}  & \multirow{1}{*}{ResNet-50} & \multirow{1}{*}{0}&\multirow{1}{*}{0}&\multirow{1}{*}{0}&\multirow{1}{*}{73.7}\\ \hline
		\multirow{1}{*}{DCP}  &\multirow{1}{*}{ResNet-50} &\multirow{1}{*}{0.5}& \multirow{1}{*}{50\%}&\multirow{1}{*}{50\%}&\multirow{1}{*}{72.4}\\ \hline
		\multirow{1}{*}{ThiNet}  &\multirow{1}{*}{ResNet-50} &\multirow{1}{*}{0.5}& \multirow{1}{*}{50\%}&\multirow{1}{*}{50\%}&\multirow{1}{*}{72.2}\\ \hline
		\multirow{1}{*}{LCP(our)} & \multirow{1}{*}{ResNet-50} &\multirow{1}{*}{0.5}&\multirow{1}{*}{50\%}&\multirow{1}{*}{50\%}&\multirow{1}{*}{\textbf{73.3}}\\ \hline
		\multirow{1}{*}{DCP}  &\multirow{1}{*}{ResNet-50} &\multirow{1}{*}{0.7}& \multirow{1}{*}{70\%}&\multirow{1}{*}{70\%}&\multirow{1}{*}{70.2}\\ \hline
		\multirow{1}{*}{ThiNet}  &\multirow{1}{*}{ResNet-50} &\multirow{1}{*}{0.7}& \multirow{1}{*}{70\%}&\multirow{1}{*}{70\%}&\multirow{1}{*}{70.8}\\ \hline
		\multirow{1}{*}{LCP(our)} & \multirow{1}{*}{ResNet-50} &\multirow{1}{*}{0.7}&\multirow{1}{*}{70\%}&\multirow{1}{*}{70\%}&\multirow{1}{*}{\textbf{71.7}}\\ \hline

	\end{tabular}
\end{table}

\begin{table}[t]
	\centering
	\caption{The pruning results on MS COCO2017. The backbone is ResNet-50, We conduct channel pruning from conv2-x to conv4-x with compression ratio 0.7. Small, medium, large are the size of objects.} \label{COCO resnet}
	\setlength{\tabcolsep}{1.5mm}
	\small
	\begin{tabular}{c|c|c|c|c|c|c}
		\hline
		\multirow{1}{*}{Method}  & \multirow{1}{*}{small} & \multirow{1}{*}{medium}& \multirow{1}{*}{large}& \multirow{1}{*}{$AP_{50}$}& \multirow{1}{*}{$AP_{75}$}& \multirow{1}{*}{mAP}\\ \hline
		\multirow{1}{*}{Original}  & \multirow{1}{*}{4.2} & \multirow{1}{*}{22.5}&\multirow{1}{*}{39.0}&\multirow{1}{*}{37.3}&\multirow{1}{*}{22.7}&\multirow{1}{*}{21.9}\\ \hline
		\multirow{1}{*}{DCP}  &\multirow{1}{*}{2.8} &\multirow{1}{*}{17.2}& \multirow{1}{*}{33.0}&\multirow{1}{*}{31.8}&\multirow{1}{*}{17.8}&\multirow{1}{*}{17.8}\\ \hline
		\multirow{1}{*}{LCP} & \multirow{1}{*}{\textbf{4.1}} &\multirow{1}{*}{20.4}&\multirow{1}{*}{38.2}&\multirow{1}{*}{35.5}&\multirow{1}{*}{\textbf{21.6}}&\multirow{1}{*}{\textbf{20.9}}\\ \hline
		\multirow{1}{*}{Relative improv.\%} & \multirow{1}{*}{\textbf{46.4}} &\multirow{1}{*}{18.6}&\multirow{1}{*}{15.8}&\multirow{1}{*}{11.6}&\multirow{1}{*}{\textbf{21.3}}&\multirow{1}{*}{\textbf{17.4}}\\ \hline
	\end{tabular}
\end{table}

\begin{table}[t]
	\centering
	\caption{The pruning results on COCO. We conduct channel pruning from conv2-x to conv4-x.} \label{COCO}
	\setlength{\tabcolsep}{1.5mm}
	\small
	\begin{tabular}{c|c|c|c|c|c}
		\hline
		\multirow{1}{*}{Method}  & \multirow{1}{*}{backbone} & \multirow{1}{*}{$\eta$}& \multirow{1}{*}{flops$\downarrow$}& \multirow{1}{*}{params$\downarrow$}& \multirow{1}{*}{mAP}\\ \hline
		\multirow{1}{*}{Original}  & \multirow{1}{*}{ResNet-50} & \multirow{1}{*}{0}&\multirow{1}{*}{0}&\multirow{1}{*}{0}&\multirow{1}{*}{21.9}\\ \hline
		\multirow{1}{*}{DCP}  &\multirow{1}{*}{ResNet-50} &\multirow{1}{*}{0.5}& \multirow{1}{*}{50\%}&\multirow{1}{*}{50\%}&\multirow{1}{*}{21.2}\\ \hline
		\multirow{1}{*}{ThiNet}  &\multirow{1}{*}{ResNet-50} &\multirow{1}{*}{0.5}& \multirow{1}{*}{50\%}&\multirow{1}{*}{50\%}&\multirow{1}{*}{22.6}\\ \hline
		\multirow{1}{*}{LCP(our)} & \multirow{1}{*}{ResNet-50} &\multirow{1}{*}{0.5}&\multirow{1}{*}{50\%}&\multirow{1}{*}{50\%}&\multirow{1}{*}{\textbf{23.1}}\\ \hline
		\multirow{1}{*}{DCP}  &\multirow{1}{*}{ResNet-50} &\multirow{1}{*}{0.7}& \multirow{1}{*}{70\%}&\multirow{1}{*}{70\%}&\multirow{1}{*}{17.8}\\ \hline
		\multirow{1}{*}{ThiNet}  &\multirow{1}{*}{ResNet-50} &\multirow{1}{*}{0.7}& \multirow{1}{*}{70\%}&\multirow{1}{*}{70\%}&\multirow{1}{*}{20.2}\\ \hline
		\multirow{1}{*}{LCP(our)} & \multirow{1}{*}{ResNet-50} &\multirow{1}{*}{0.7}&\multirow{1}{*}{70\%}&\multirow{1}{*}{70\%}&\multirow{1}{*}{\textbf{20.9}}\\ \hline

	\end{tabular}
\end{table}

\begin{table}[t]
	\centering
	\caption{The pruning results on PASCAL VOC2007. We conduct channel pruning from conv2-x to conv4-x. CR means Contextual RoIAlign.} \label{ablation}
	\setlength{\tabcolsep}{1.5mm}
	\small
	
	\begin{tabular}{c|c|c|c|c}

		\hline
		\multirow{1}{*}{Method}  & \multirow{1}{*}{backbone} & \multirow{1}{*}{flops$\downarrow$}& \multirow{1}{*}{params$\downarrow$}& \multirow{1}{*}{mAP}\\ \hline

		\multirow{1}{*}{DCP}  &\multirow{1}{*}{ResNet-50} & \multirow{1}{*}{70\%}&\multirow{1}{*}{70\%}&\multirow{1}{*}{70.2}\\ \hline
		
		\multirow{1}{*}{LCP+RoIAlign} & \multirow{1}{*}{ResNet-50} &\multirow{1}{*}{70\%}&\multirow{1}{*}{70\%}&\multirow{1}{*}{\textbf{71.1}}\\ \hline
		\multirow{1}{*}{LCP+CR}  &\multirow{1}{*}{ResNet-50} & \multirow{1}{*}{70\%}&\multirow{1}{*}{70\%}&\multirow{1}{*}{\textbf{71.7}}\\ \hline
	\end{tabular}
\end{table}
\begin{table}[t]
	\centering
	\caption{The pruning results on PASCAL VOC2007. We conduct channel pruning from conv1-1 to conv5-3 .} \label{loss}
	\setlength{\tabcolsep}{1.5mm}
	\small
	\begin{tabular}{c|c|c|c|c|c}
		\hline
		\multirow{1}{*}{Method}  & \multirow{1}{*}{backbone} & \multirow{1}{*}{$\eta$}& \multirow{1}{*}{flops$\downarrow$}& \multirow{1}{*}{params$\downarrow$}& \multirow{1}{*}{mAP}\\ \hline
		\multirow{1}{*}{Original}  & \multirow{1}{*}{VGG-16} & \multirow{1}{*}{0}&\multirow{1}{*}{0}&\multirow{1}{*}{0}&\multirow{1}{*}{77.4}\\ \hline
		\multirow{1}{*}{$\mathcal{L}_{re}$+$\mathcal{L}_{ac}$}  &\multirow{1}{*}{VGG-16} &\multirow{1}{*}{0.75}& \multirow{1}{*}{75\%}&\multirow{1}{*}{75\%}&\multirow{1}{*}{74.7}\\ \hline
		\multirow{1}{*}{$\mathcal{L}_{re}$+$\mathcal{L}_{ac}$+$\mathcal{L}_{ar}$} & \multirow{1}{*}{VGG-16} &\multirow{1}{*}{0.75}&\multirow{1}{*}{75\%}&\multirow{1}{*}{75\%}&\multirow{1}{*}{\textbf{75.2}}\\ \hline
	\end{tabular}
\end{table}
\begin{figure*}[t]
	\begin{center}
		\includegraphics[width=1\linewidth]{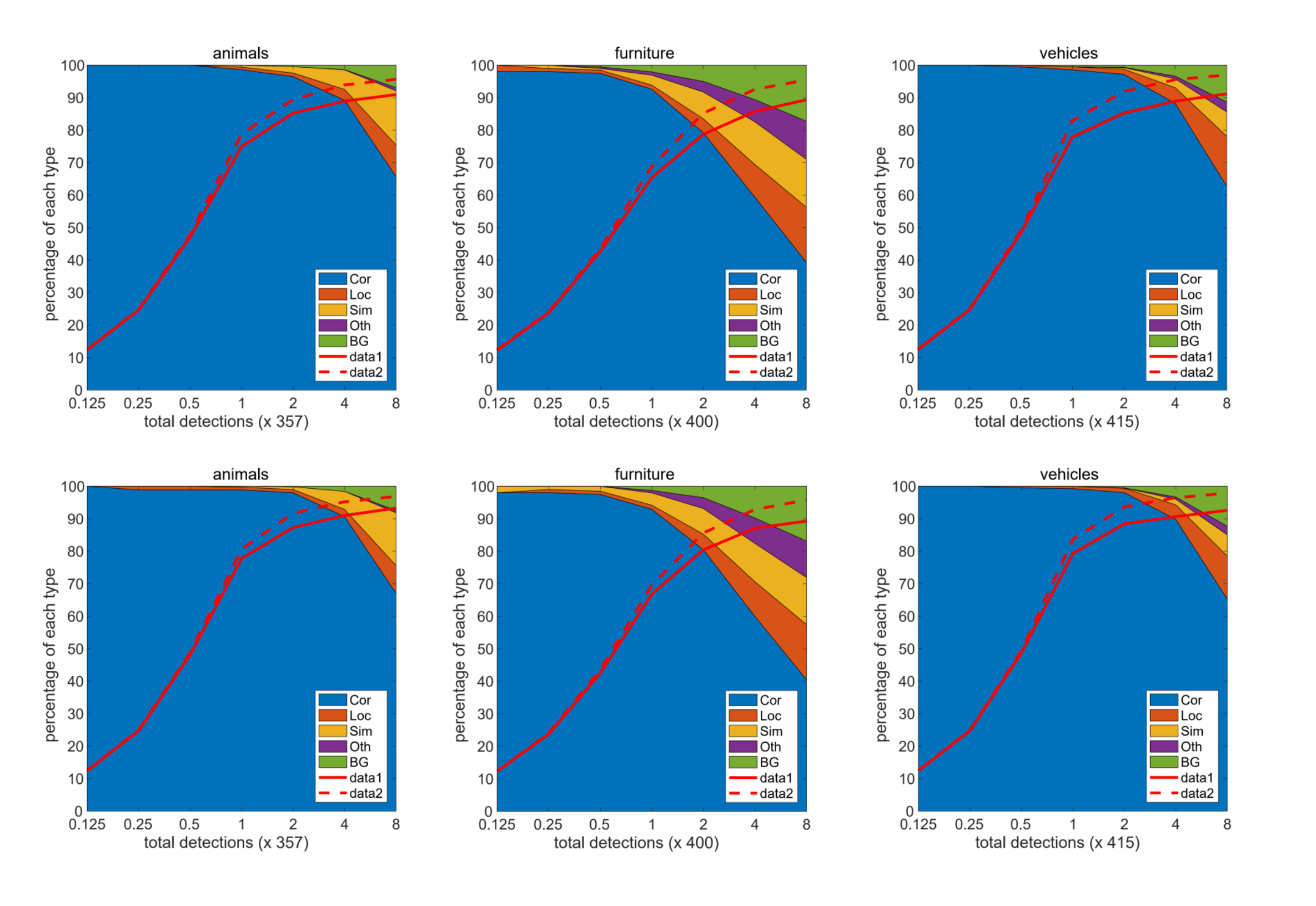}
	\end{center}
	\caption{Visualization of the performance of ThiNet (top row) and our method (bottom row) on animals, furniture, and vehicles classes in the VOC 2007 test set. The figures show the cumulative fraction of detections that are correct (Cor) or false positives due to poor localization (Loc), confusion with similar categories (Sim), with others (Oth), or with background (BG). The solid red line reflects the change of recall with the ‘strong’ criteria (0.5 jaccard overlap) as the number of detections increases. The dashed red line uses the ‘weak’ criteria (0.1 jaccard overlap).}
	\label{ana}
\end{figure*}
\begin{figure*}[t]
	\begin{center}
		\includegraphics[width=1\linewidth]{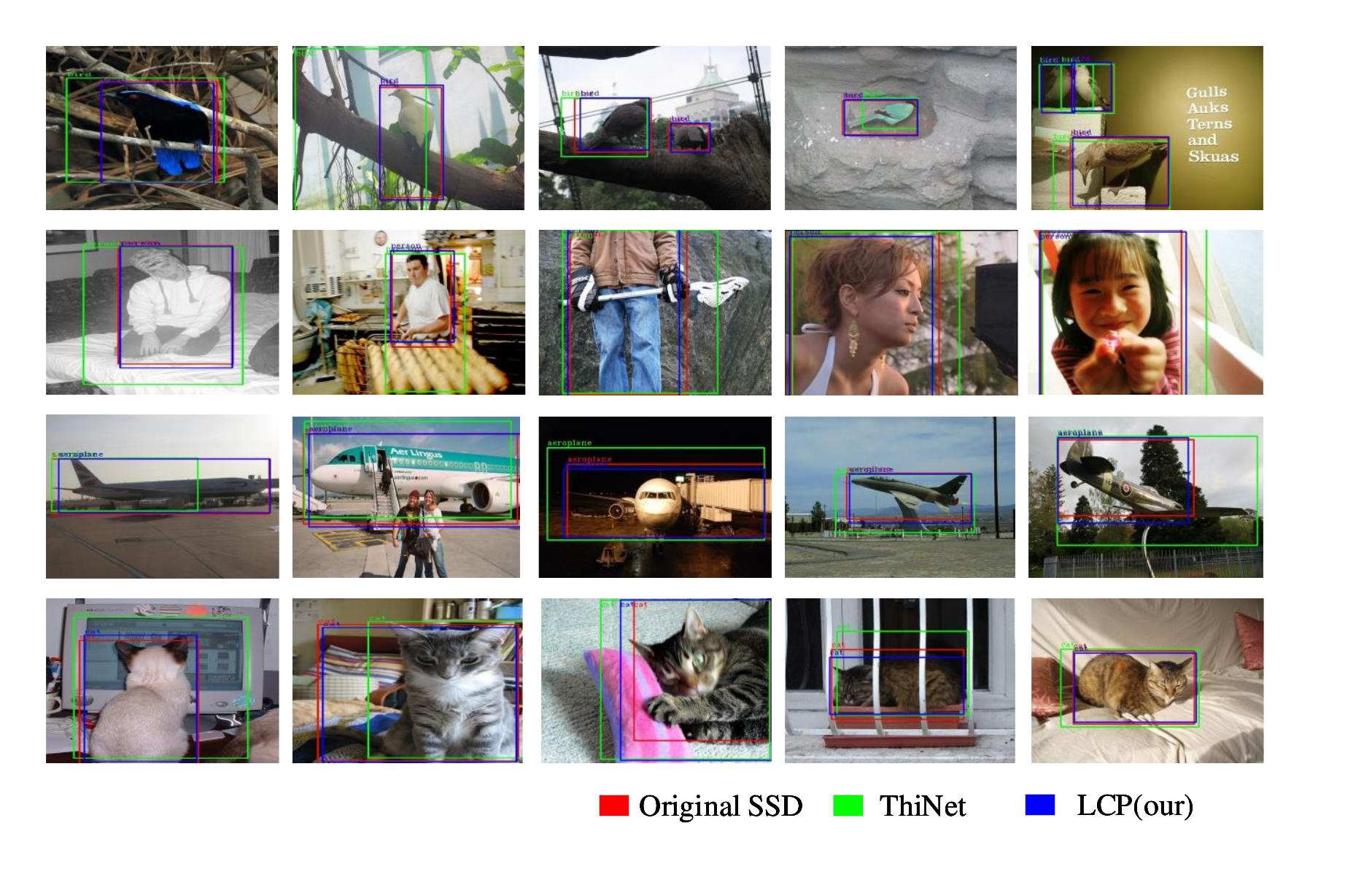}
	\end{center}
	\caption{The predictions of original SSD, models pruned by Thinet and our LCP. We prune the VGG-16 by 75\% on PASCAL VOC.}
	\label{visual}
\end{figure*}

\subsection{Experiments on MS COCO}
In this section, we prune the ResNet-50 by 70\% on COCO2017. We report the results in Tab. \ref{COCO resnet} and Tab. \ref{COCO}. From the results, our method achieves a better performance than the DCP and ThiNet, which further illustrates the effectiveness of our approach. It is noted that compared with DCP, LCP has larger gain on small objects. In addition, the higher the IoU threshold, the greater improvement of our method. This indicates that our method retains more localization information and can obtain more accurate predictions.\\
\subsection{Ablation Analysis}
\textbf{Gradient Analysis.} In this section, we prune the VGG-16 from conv1-1 to conv5-3 with compression ratio 0.75 On PASCAL VOC. Then we count the percentage of the gradients generated by the three losses during the pruning process. From Fig. \ref{gradient}, we see that the gradient of regression loss play a important role during the pruning process, which proves that the localization information is necessary. The gradient generated by reconstruction error only works in the shallow layers while the localization-aware loss contributes to the channel pruning process each layer.\\
\textbf{Component Analysis.} In this section, in order to verify the effectiveness of the two points we propose, we prune the SSD based on ResNet-50 by 70\% with different combinations of our points. We report the results in Tab.\ref{ablation}. From the results, we can get that each part of the method we propose contributes to the performance.\\
\textbf{Loss Analysis.} In order to explore the importance of the gradient of regression loss, we prune the SSD based on VGG-16 by 75\% with different losses. We report the results in Tab. \ref{loss}. From the results, we can know that the performance of our method drops a lot without the gradient of the regression loss during the pruning stage, which shows that the regression branch contains important localization information. 
\subsection{Qualitative Results}
\textcolor[rgb]{0,0,1}{To demonstrate the effectiveness of our proposed method in details, we use the detection analysis tool from \cite{hoiem2012diagnosing}. Figure 7 shows that our model can detect various object categories with high quality (large blue area). The recall is higher than 90\%, and is much higher with the ‘weak’ (0.1 jaccard overlap) criteria. We can also observe that comparing with ThiNet, Our method has larger correct area, which indicates our superior performance.}
\subsection{Visualization of predictions}
In this section, we prune the SSD based on VGG-16 by 75\%  and we compare the original model with the pruned models. From Fig. \ref{visual}, we can find that the predictions of our method are closed to the predictions of the original model while the predictions of ThiNet are far away. It is proved that our method reserve more localization information for bounding box regression.
\section{Conclusions}
In this paper, we propose a localization-aware auxiliary network which allows us to conduct channel pruning directly for object detection. First, we design the auxiliary network with a contextual \textcolor[rgb]{0,0,1}{RoIAlign} layer which can obtain precise localization information of the default boxes by pixel alignment and enlarges the receptive fields of the default boxes when pruning shallow layers. Then, we construct a loss function for object detection task which tends to keep the channels that contain the key information for classification and regression. Visualization shows our method reserves layers with more localization information. Moreover, extensive experiments demonstrate the effectiveness of our method.

\section{Acknowledge}
This work was supported by the National Natural Science Foundation of China under Grants 91748204, 61976227 and 61772213 and in part by the Wuhan Science and Technology Plan under Grant 2017010201010121 and Shenzhen Science and Technology Plan under Grant JCYJ20170818165917438.

\bibliography{reference}

\end{document}